\documentclass[journal]{IEEEtran}
\usepackage{preamble}
\usepackage{flushend}
\ifCLASSINFOpdf
\else
   \usepackage[dvips]{graphicx}
\fi
\usepackage{url}

\usepackage{graphicx}

\begin{document}

\title{Ensemble Neural Representation Networks}

\author{Milad Soltany Kadarvish$^{\ast}$, Hesam Mojtahedi$^{\ast}$, Hossein Entezari Zarch$^{\ast}$, Amirhossein Kazerouni$^{\ast}$,\\ Alireza Morsali, Azra Abtahi, and Farokh Marvasti

\thanks{Star indicates equal contributions.}
\thanks{M. S. Kadarvish and A. Kazerouni are with the School of Electrical Engineering, Iran University of Science and Technology, Tehran, Iran (e-mail: soltany.m.99@gmail.com; amirhossein477@gmail.com).}
\thanks{H. Mojtahedi and H. E. Zarch are with the School of Electrical and Computer Engineering, College of Engineering University of Tehran, Tehran, Iran (e-mail: hsm.moj@gmail.com; h.entezari78@gmail.com).}
\thanks{A. Morsali is with the Department of Electrical and Computer Engineering, McGill University, Montreal, Canada (e-mail: alireza.morsali@mail.mcgill.ca).}
\thanks{A. Abtahi is with the Department of Electrical and Information Technology, Lund University (e-mail: azra.abtahi\_fahliani@eit.lth.se)}
\thanks{F. Marvasti is with the Department of Electrical Engineering, Sharif University of Technology, Tehran, Iran (e-mail: marvasti@sharif.edu).}}

\markboth{}
{Shell \MakeLowercase{\textit{et al.}}: Bare Demo of IEEEtran.cls for IEEE Journals}
\maketitle

\begin{abstract}
Implicit Neural Representation (INR) has recently attracted considerable attention for continuous characterization various types of signals. The existing INR techniques require lengthy training processes and high-performance computing. In this letter, we propose a novel ensemble architecture for INR that resolves the aforementioned problems. In this architecture, the representation task is divided into several sub-tasks done by independent sub-networks. We show that the performance of the proposed ensemble INR architecture may decrease if the dimensions of sub-networks increase. Consequently, it is vital to suggest an optimization algorithm to find suitable structures for the ensemble networks, which is also done in this paper. According to the simulation results, the proposed architecture not only has significantly fewer floating-point operations (FLOPs) and less training time, but it also has better performance in terms of Peak Signal to Noise Ratio (PSNR) compared to those of its counterparts. {\scriptsize (The source code is available at
\url{https://github.com/AlirezaMorsali/ENRP})}

\end{abstract}


\IEEEpeerreviewmaketitle

\section{Introduction}

\IEEEPARstart{R}{ecently},
implicit neural representation (INR) has been introduced as a new way of representing complex signals, while such signals are conventionally expressed explicitly, e.g., colored images and audio signals as 3D and 1D arrays, respectively. INR takes advantage of the universality of artificial neural networks, e.g., a Multi-Layer Perceptron (MLP), to approximate the underlying generator function of the given signal. For instance, instead of storing the pixel values of a grayscale image in a matrix, pixel coordinates are given as input to an MLP, and the output is the pixel values. INR is particularly useful in computer graphics, specifically in encoding images \cite{radford2015unsupervised, compositional} and reconstructing 3D shapes by either occupancy networks \cite{Occupancy, Generative} or signed distance functions \cite{Deepsdf, Sal, ImplicitSurfaceRep, peng2020convolutional}.

Rahaman \emph{et al.} in \cite{rahaman2019spectral} have mathematically proved that conventional ReLU Deep Neural Networks (DNN) are deficient in representing high-frequency components of signals. This phenomenon is known as spectral bias in piece-wise linear networks. Recently, alternative network architectures have been proposed which can successfully represent complex signals such as images and 3D shapes \cite{Siren, Fourier_features, NeRF}. Furthermore, Tancik \emph{et al.} in \cite{Fourier_features} investigated that the tendency of piece-wise linear networks (i.e., ReLU networks) is towards training low-frequency features first. Consequently, they introduced the Fourier feature network that maps the input to a higher dimension space using sinusoidal kernels to facilitate the learning of high-frequency data. In this respect, Mildenhall \emph{et al.} in\cite{NeRF} utilize the Fourier feature network to represent continuous scenes as 5D neural radiance fields (NeRFs). As a concurrent work with \cite{Fourier_features}, Sitzmann \emph{et al.} in \cite{Siren}, introduced sinusoidal representation networks or SIRENs as a novel approach to represent various complicated signals. They replaced the ReLU activation function with a periodic one and exploited a unique and controlled initialization scheme.

While it has been shown that INR networks outperform conventional methods in several tasks \cite{Siren, NeRF, NeuralVolumes}, they are still far from being widely used. Specifically, the existing INR networks require lengthy training processes and extensive computational resources. Consequently, aside from improving the performance of INR, these networks must be optimized to be faster and more computationally efficient. One way of achieving this goal is using local representation techniques. Local methods have been used extensively for processing complex signals. For instance, 3D models use these methods to enhance the accuracy and reduce the computational cost in representing 3D scenes, such as recent voxelized implicit models \cite{DeepVoxels,jiang2020local, LDIF, DeepLocalShapes}. % While NeRF \cite{NeRF} utilizes one ReLU MLP network, Reiser in \emph{et al.} \cite{KiloNeRF} introduce the KiloNeRF model that represents scenes with a large number of tiny and independent ReLU MLPs by subdividing each scene into a 3D grid.
In \cite{Modulated}, Mehta \emph{et al.} also decomposed 2D images into sets of regular grid cells, each represented by a latent code. Then, each of these latent codes is fed into a modulation network, and each image cell enters a synthesis network.

% \color{red}
This paper proposes a novel architecture that uses an ensemble of networks to split the representation task into several smaller sub-tasks. By doing so, each sub-network pays attention to the local information of the signal and is trained only on a specified segment of the signal.
% Although KiloNeRF utilizes ReLU MLPs as its sub-networks, we use SIRENs.
We have shown that the performance of the proposed ensemble structure does not always follow an upward trend when the number of the sub-networks increases or when wider sub-networks are used.
% Hence, proposing an algorithm for the sub-optimal design of the sub-networks is of utmost importance, which we have also presented in this paper.
Thus, we present an algorithmic approach for designing the sub-networks.
Exploiting the proposed algorithm, the ensemble network not only can achieve better performance compared to its counterparts but also the number of FLOPs decreases significantly. \color{black} In addition, our novel approach has the generality to apply to different INR structures, such as ReLU-based MLPs, SIREN, Fourier feature networks, etc., while improving the results considerably.

\begin{figure*}[t]
    % \vspace{-20px}
	\centering
    \includegraphics[width=0.85\textwidth]{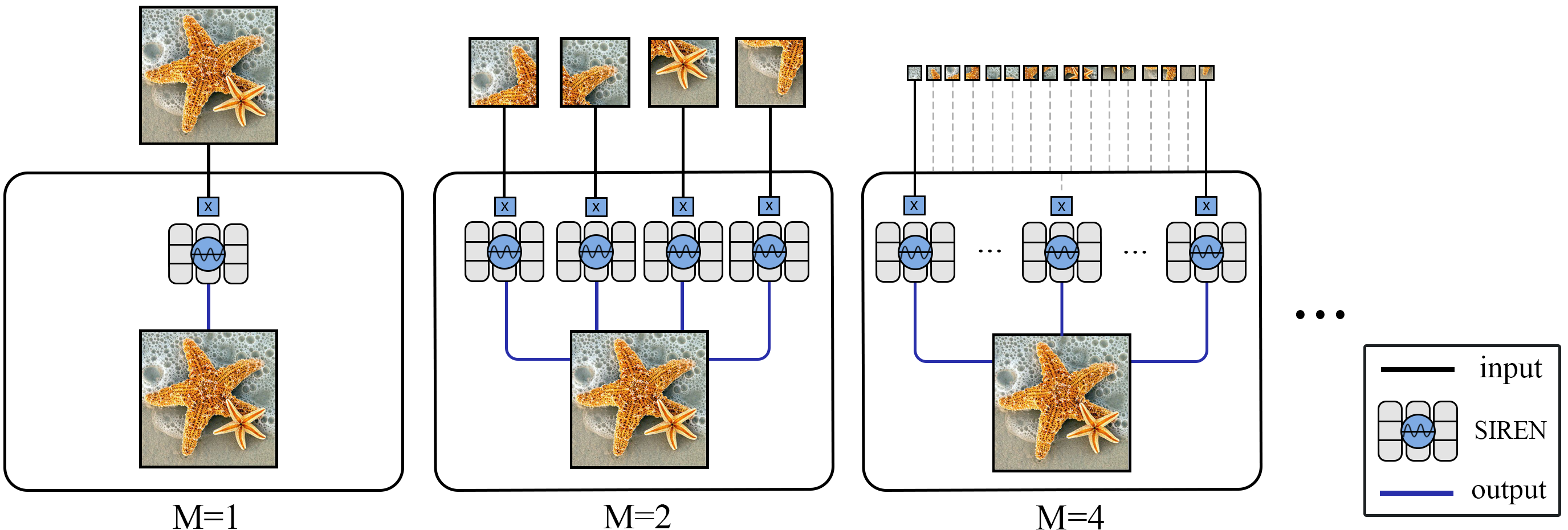}
    \vspace{-5px}
    \caption{\textbf{Overview of our approach:} After dividing the input image into cells of a $M\times M$ grid, each image cell is fed into an individual network. The final output is generated by aggregating all outputs of sub-networks.} \color{black}
    \label{fig:Architecture}
    \vspace{-5px}
    % \vspace{-1em}
\end{figure*}

%%%%%%%%%%%%%%%%%%%%%%%%%%%%%%%%%%%%%%%%%%%%%%%%%
%%%%%%%%%%%%%%%%%%%%%%%%%%%%%%%%%%%%%%%%%%%%%%%%%
%%%%%%%%%%%%%%%%%%%%%%%%%%%%%%%%%%%%%%%%%%%%%%%%%
%%%%%%%%%%%%%%%%%%%%%%%%%%%%%%%%%%%%%%%%%%%%%%%%%

% \section{Guidelines For Manuscript Preparation}

\color{black}
\section{Problem Formulation}
% \vspace{-15px}
% \label{sec:formulation}
% \vspace{-2mm}
Without the loss of generality, in this paper, we focus on INR for colored images.
However, our proposed method can be easily generalized to multi-dimensional signals as well. For a given $N\times N$ colored image, i.e., $S\in\R^{N\times N \times 3}$, the goal of INR is to find $\Phi: \R^2\to\R^3$ such that
\begin{equation}
  \label{eq:1}
     \norm{\Phi\big{(}x_1^r, x_2^c\big{)} - S[r, c, :]}= 0
\end{equation}
for $r=1,2,\dots,N$, $c=1,2,\dots,N$ where $x_1^r=2(r-1/(N-1))-1$, $x_2^c=2(c-1/(N-1))-1$, and $S[r, c, :]$ denotes all the entries of $S$ on the $r^{th}$ row and the $c^{th}$ column.
Existing works in the literature \cite{Siren,Fourier_features} use an MLP to approximate $\Phi(x_1, x_2)$ and train the MLP using the dataset $\pazocal{D}=\{(\small{(x_1^r, x_2^c)}, S[r, c, :])\}_{r,c=1}^{N, N}$.
% where $i=1,2,\dots, N_hN_w$.
\vspace{5px}
However, in \eqref{eq:1}, $\Phi(\cdot)$ is a generic mapping that can be further expressed as the sum of several sub-functions to reduce computational complexity.
In particular, we can write
\begin{equation}
\label{eq:2}
  \Phi(x_1, x_2)=\sum_{m=1}^{M^2}\phi_m(x_1, x_2)I_m(x_1, x_2)
\end{equation}
% \fxwarning{Fix Hessam and Hossein}
where $\phi_m(x_1, x_2)$'s are continuous functions from $\R^2$ to $\R^3$ and $I_m(x_1, x_2)$'s are indicator functions, i.e.,
\begin{equation}
\label{eq:3}
    I_m(x_1,x_2)=
    \begin{cases}
1 &(x_1,x_2)\in D_m\\
0 &(x_1,x_2)\notin D_m
\end{cases}
\end{equation}
where $D_m=$
\begin{equation}
\label{eq:4}
\small
\{(x_1,x_2)|2\frac{i-1}{M}-1<x_1<\frac{2i}{M}-1,2\frac{j-1}{M}-1<x_2<\frac{2j}{M}-1\}
\end{equation}
with $i=\lfloor\frac{m-1}{M}\rfloor+1$ and $j=\textrm{mod}(m-1,M)+1$.

% \vspace{10px}
In what follows, we focus on a structured ensemble architecture for $\Phi(x_1, x_2)$ and optimize its hyperparameters.
\begin{table}[t]
% \vspace{-3em}
    \centering
    \label{algorithm:main}
    \begin{tabular}{l}
        \toprule
        \textbf{Algorithm 1} The Proposed Iterative Method\\
        \midrule
        \textbf{Inputs:} Maximum iteration number (${iter}_{max}$),\\
	   % Search set for the depth of the MLPs  ($D_1$),\\
	   % Search set for the width of the MLPs  ($W_1$),\\
	    Search set for the depth of the MLPs ($D_1$=  $\left \{ 1,2,3,4,5 \right \}$),\\
	    Search set for the width of the MLPs ($W_1$=  $\left \{ 16,32,64,\dots \right \}$),\\
		The number of parallel MLPs ($M$),\\
    	The maximum number of FLOPs (${F}_{max}$),\\
        Loss function($L_{M, d, w}$).\\
        \textbf{Initialization:} $w_1 = 256$,\quad $d_1 = 3$ \\
        % \quad \quad \quad \quad M is Constant\\
        \textbf{Outputs:} The optimal depth and width for the MLPs ($d$ \textrm{and} $w$)\\
%         Initialize the depth and the width of the MLPs according to the\\
%         proposed MLP architecture for SIREN: $d_0=..., w_0=...$\\
        1.\textbf{for}{\; $i =1:{iter}_{max}$} \textbf{do}\\
        % 2.\; \quad \quad  $d_i=\underset{\hat{d}}{\arg\min} \quad{ L_m(\hat{d},w_{i-1})}$,\\
        % 2.\; \quad \quad $w_{i+1}=\underset{\hat{w}}{\arg\min} \quad{ L_{M, d_i, \hat{w}}(S)}$\\
        2.\quad $w_{i+1}=\underset{\hat{w}}{\arg\max}\quad{ 20 \log_{10}(\frac{1}{L_{M, d_i, \hat{w}}(S)}}) - \alpha \log_{10}(FC\footnote{FLOP count})$\\
            \quad \quad \quad \quad s.t.\quad  $FC \leq F_{max}$\\
        % 3.\; \quad \quad  $d_{i+1}=\underset{\hat{d}}{\arg\min} \quad{     L_{M, \hat{d}, w_{i+1}}(S)}$,\\
        3.\quad $d_{i+1}=\underset{\hat{d}}{\arg\max}\quad{     20\log_{10}(\frac{1}{L_{M, \hat{d}, w_{i+1}}(S)}}) - \alpha \log_{10}(FC)$\\
            \quad \quad \quad \quad s.t.\quad  $FC \leq F_{max}$ \\
        4.\quad \textbf{if} {($w_{i+1}==w_{i}$ \& $d_{i+1}==d_{i}$) } \textbf{then} \\
        5.\quad \quad \quad $d \gets {d}_i$, $w \gets {w}_i$\\
        6. \quad \quad \quad break\\
        7.$d \gets {d}_{{iter}_{max}}$\\
        8.$w \gets {w}_{{iter}_{max}}$\\
        % \textbf{Note:} The output of the \hp, i.e., $\xth$, is identical to $\xtd$.\\
        \bottomrule
    \end{tabular}
    % \vspace{-1.5em}
%     \begin{tablenotes}
%      \item \begin{FlushLeft}^1\textrm{FLOP Count} \end{FlushLeft}
%   \end{tablenotes}
\vspace{-10px}
\end{table}
\section{The Proposed Ensemble INR Network}
\label{sec:INR}
In this section, we first present our network architecture for the ensemble network. Then, we discuss the importance of the sub-network design in the proposed architecture. We take the SIREN structure to present our approach. Further, we show the proposed ensemble network results for different INR structures.

\subsection{Network Architecture}

The model $\Phi(x_1, x_2)$ consists of $M^2$ sub-networks each denoted by $\phi_m(x_1, x_2)$, which is responsible to learn a cell of image. Therefore, the image is uniformly divided into $M\times M$ sub-images denoted by $S_{i,j}$.

We consider a similar network for all $\phi_m$'s to keep the ensemble network symmetric.
Furthermore, we can observe from \eqref{eq:1} and \eqref{eq:2}, for $m=1,2,\dots,M^2$, that the function $\phi_m$ is only defined on the interval $D_m$ defined in \eqref{eq:4}. Consequently, to simplify the hyper-parameter tuning, a simple positional encoding is first performed on the input of each sub-network $\phi_m$ such that $D_m$ maps to ${([-1,1],[-1,1])}$.

The objective function of the m-th sub-network can be written as follows
% \vspace{-5px}
\begin{equation}
    \norm{\phi_m(x_1^r, x_2^c) - S_{i,j}[r, c, :]}=0
\end{equation}
% \vspace{-5px}
where $i=1,2,\dots,M$, and $j=1,2,\dots,M$ determine the position of the m-th cell in the image grid.

% \color{red}
As it is depicted in Fig. \ref{fig:Architecture}, our proposed method divides the given image into several parts, then each part is fed into an individual sub-model. Fig.\ref{fig:Architecture} exemplifies the use of our method for images with $M=1, 2, \textrm{and}\; 4$. Each of these ensemble networks, $\Phi(x_1, x_2$), will be trained in a parallel fashion. In other words, their sub-networks will be trained simultaneously. As shown in Fig. \ref{fig:Architecture}, for $M = 1$, a single network is utilized to represent the entire image, similar to SIREN. For $M = 2$ and $M = 4$, the image is symmetrically divided into 4 and 16 non-overlapping parts, respectively. Further, each part is represented with an individual network
%MLP with sine activation functions
. At the end of each forward pass, the outputs of these networks will be aggregated to form a complete signal. Afterward, the loss will be calculated, and the backward pass will occur. \color{black}
\begin{figure}[t]
	\centering
% % 	\advance\leftskip-0.5cm
    % \vspace{-20px}
    \includegraphics[width=8cm]{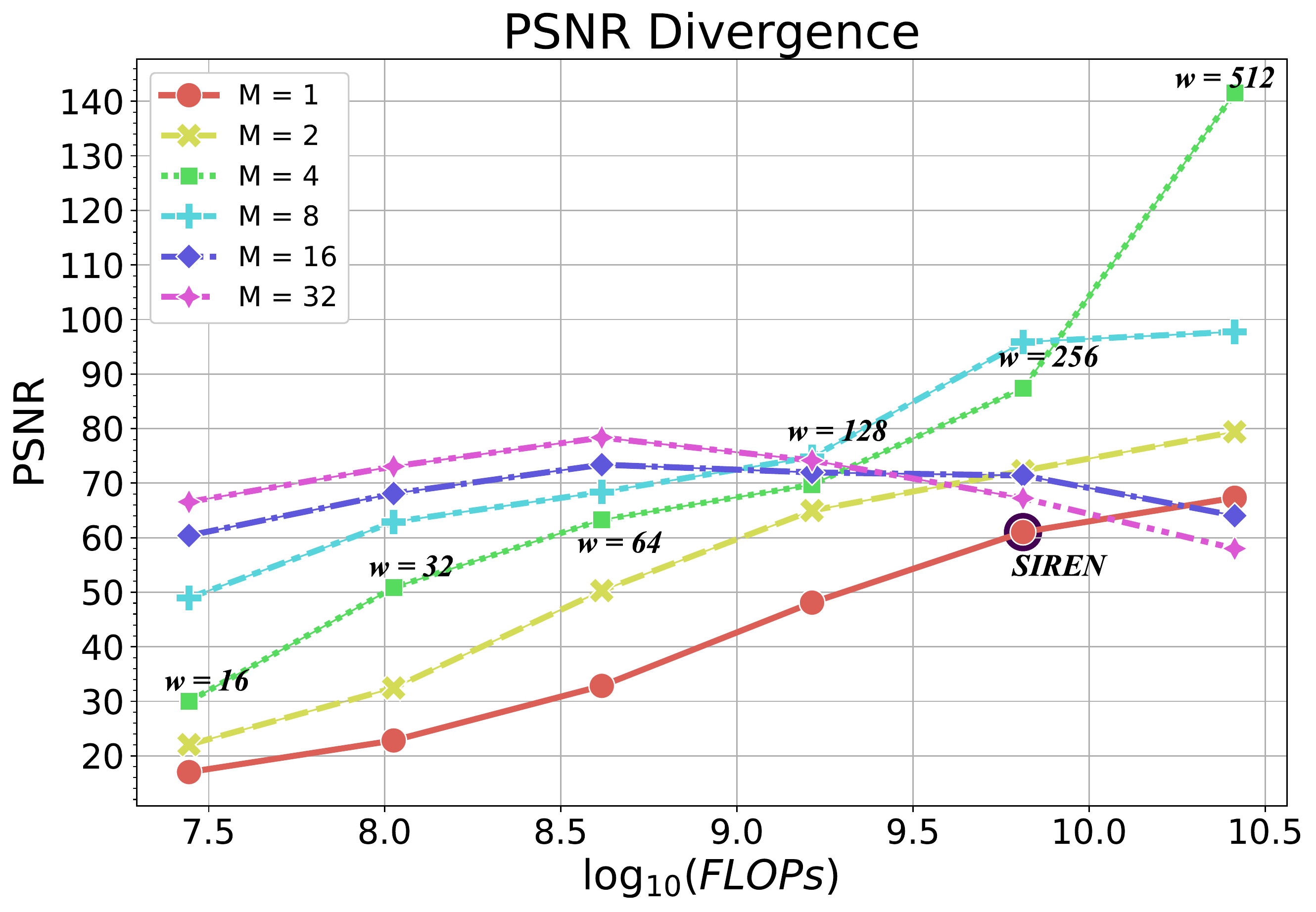}
    % \vspace{-10px}
    \vspace{-5px}
    \caption{Divergence in PSNR values of ensemble networks with different $w$ values.}
    \label{fig:divergence}
    \vspace{-1em}
    \setlength{\belowcaptionskip}{-20pt}
\end{figure}

\vspace{-1em}
\subsection{The Proposed Network PSNR Divergence}
We have evaluated different configurations of the network and observed that increasing the width of sub-networks does not always lead to a higher PSNR. In order to be certain that randomness is not behind this behavior, we have repeated the training process for 20 different images; for each image, we re-trained each network 10 times. Moreover, we calculated the average of PSNR values for each model configuration. The images were $128\times 128$ and the number of hidden layers for each sub-network, $d$, was equal to 3.
\begin{figure}[b]
	\centering
    \includegraphics[width=7cm]{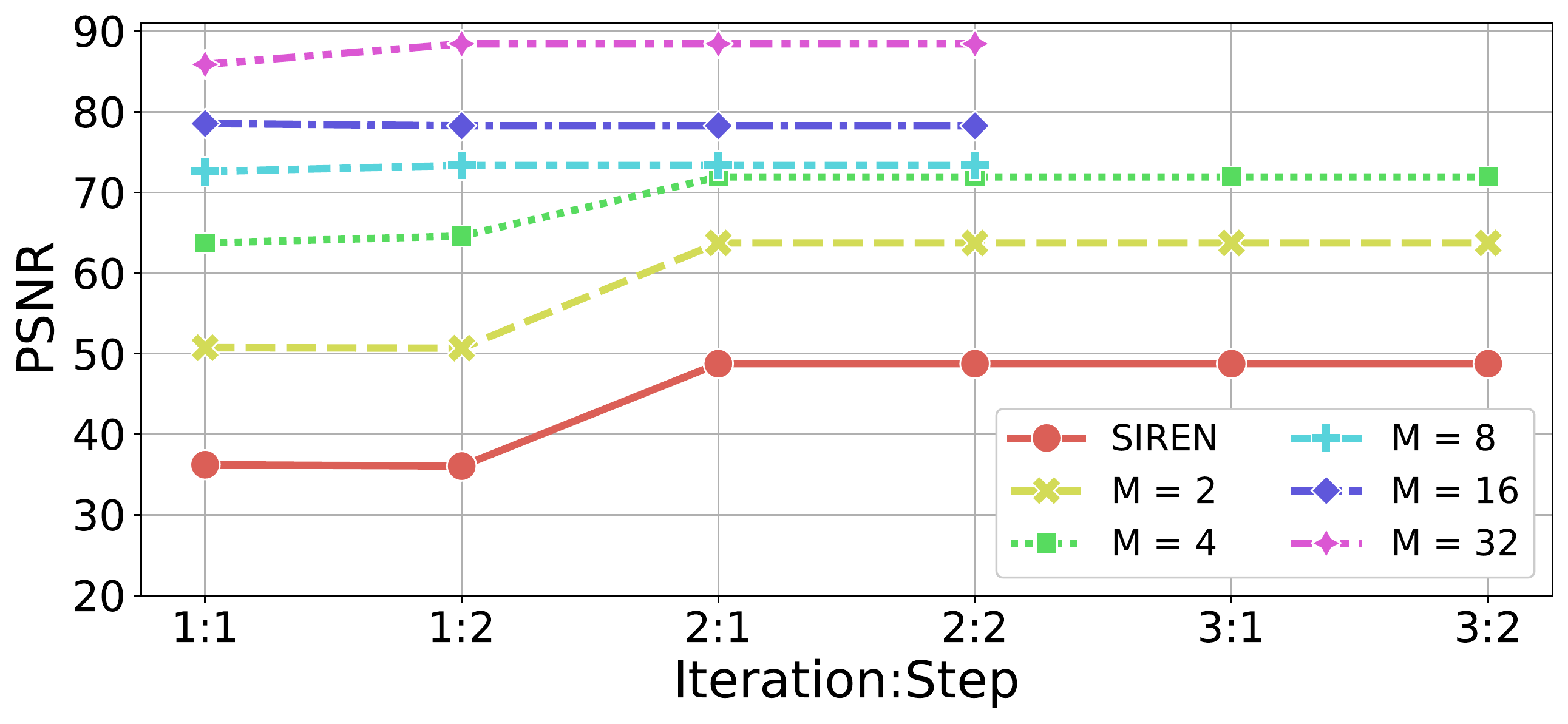}
    % \vspace{-1em}
    \vspace{-5px}
    \caption{\textbf{Algorithm Results}: PSNR versus iteration number of the proposed algorithm for $M = \{ 2, 4, \dots, 32\}$ and SIREN.}
    % \vspace{-2em}
    \label{fig:algorithm_plot}
\end{figure}
The PSNR values versus the number of FLOPs (corresponding to various widths values of the sub-networks, w) for some sub-network configurations are depicted in Fig \ref{fig:divergence}. This result shows the importance of defining and solving an optimization problem that determines the optimal network configuration. In the following section, an optimization problem is defined, and to solve it, an iterative algorithm is proposed.

\begin{figure}[t]
    % \vspace{-8px}
    % \vspace{-1.5em}
    \centering
    \includegraphics[width=8cm]{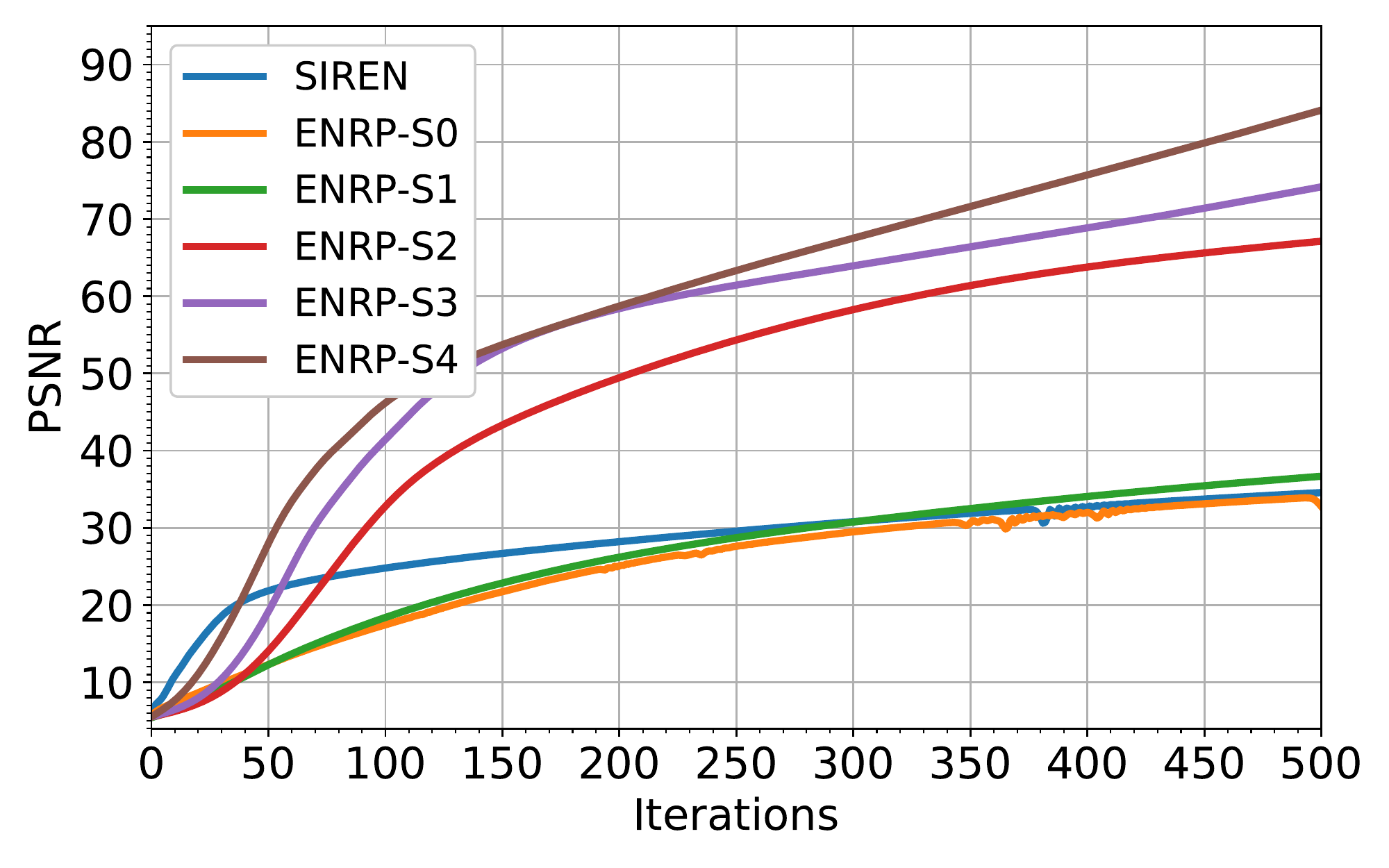}
    % \vspace{-1em}

    \vspace{-7px}
    \caption{\textbf{Convergence Rates:} PSNR versus training iteration number for ENRP models and SIREN.}
    \label{fig:result_iter}
    \vspace{-1em}
\end{figure}
\color{black}

\section{The proposed Algorithm for Designing Sub-networks}
% \vspace{-0.5em}
\label{sec:algo}
We now focus on finding the optimal ensemble INR architecture, while satisfying a constraint on a given maximum number of FLOPs.
The loss of the network for a given image $S$ can be expressed as
% \vspace{-5px}
\begin{equation}
\label{loss_function}
L_{M, w, d}(S) = \sum_{i=1}^{{M}}\sum_{j=1}^{{M}}\sum_{r=1}^{\frac{N}{M}}\sum_{c=1}^{\frac{N}{M}}\norm{\phi_m^{d,w}(x_1^r, x_2^c) - S_{i,j}[r, c, :]}
\end{equation} where, $w$ and $d$ are pivotal factors for scaling the width and depth of the network, and $m = M\times (i-1) + j$.  We can then write the optimization problem as
% \begin{equation}
%     \label{min_problem}
%     \begin{split}
%         \underset{M,w,d}{arg\min} \quad & \mathbb{E}\{L_{M}(S)\} \\
%         s.t.\quad &\\
%         &FLOPs \leqslant Maximum \; Number \; {of} \; {FLOPs}
%     \end{split}
% \end{equation}
\begin{equation}
    \label{min_problem}
    \begin{aligned}
        \underset{M, w, d}\min \quad & \mathbb{E}\{L_{M, w, d}(S)\}\\
        \textrm{s.t.} \quad & \textrm{FLOPs} \leqslant \textrm{Maximum} \; \textrm{Number} \; \textrm{of} \; \textrm{FLOPs}
    \end{aligned}
\end{equation}

Due to the difficulty of solving the architecture optimization problem presented in (\ref{min_problem}),  we need to solve it in a sub-optimal manner. Thereby, we propose the iterative search algorithm in (\hyperref[algorithm:main]{1}) that does not proceed unless there is an improvement in the resultant PSNR value.

\begin{table}[b]
\vspace{-5px}
  \begin{center}
    \caption{\textbf{Image:} The configurations, PSNRs, and FLOP counts of the proposed ensemble networks for $M = \{2, 4, \dots, 32\}$ and SIREN, after 500 training steps.}
    \label{tab:table1}
    \vspace{-10px}
    \begin{tabular}{c|c|c|c|c|c}
      \textbf{Name} & \textbf{M} & \textbf{Depth (d)} & \textbf{Width (w)} & \textbf{$\#$FLOPs} & \textbf{PSNR (dB)}\\
      \hline SIREN & 1 & 3 & 256 & 6.48 G & 34.12\\
      \hline ENRP-S0 & 2  & 5 & 64 &  681.58 M & 33.80 \\
      \hline ENRP-S1 & 4  & 2 & 64 &  278.92 M & 36.51 \\
      \hline ENRP-S2 & 8  & 1 & 128 &  557.84 M & 67.51 \\
      \hline ENRP-S3 & 16 & 1 & 128 &  557.84 M & 75.26 \\
      \hline ENRP-S4 & 32 & 1 & 128 &  557.84 M & 83.08
    \end{tabular}
  \end{center}
  \vspace{-8px}
% \vspace{-6em}
\end{table}

In order to regulate the objective function not to result in a model with a much larger FLOP count that yields a slightly higher PSNR value, the algorithm has a regulation term weighted by $\alpha$ to adjust the relative importance of FLOP count compared to the PSNR value. In our experiment, we set $\alpha$ to be equal to 7.

% \color{red}
The initial values for $d$ and $w$ are set to be equal to those of SIREN, i.e., $w_1=256$ and $d_1=3$. \color{black} For a given M, in each iteration, there are two steps. In the first step, the value of $\hat{w}$ is chosen to have the best performance by satisfying the FLOPs constraint. In the next step, after replacing the acquired $\hat{w}$, the algorithm chooses the best $d$ to reach the highest PSNR. The algorithm comes to a stop when either the maximum number of iterations, $iter_{max}$, is reached, or the results produced in the last step show no progress compared to those of the one before it.

%%%%%%%%%%%%%%%%%%%%%%%%%%%%%%%%%%%%%%%%%%%%%%%%%%%%%%%%%%%%%%%%%%%%%%%%%%%%%%%%%%%%%%%%%%%%%%%%%%%%%%%%%%%%%%%%%%%%%%%%%%%%%%%%%%%%%%%%%%%%%%%%%%%%%%%%%%%%%%%%%%%%%%%%%%%%%%%%%%%%%%%%%%%%%%%%%%%%%%%%%%%%%%%%

% \vspace{-4em}

\section{Simulation Results}
\label{sec:result}
\begin{figure*}[htp]
    % \vspace{-20px}

	\centering
    \includegraphics[width=0.85\textwidth]{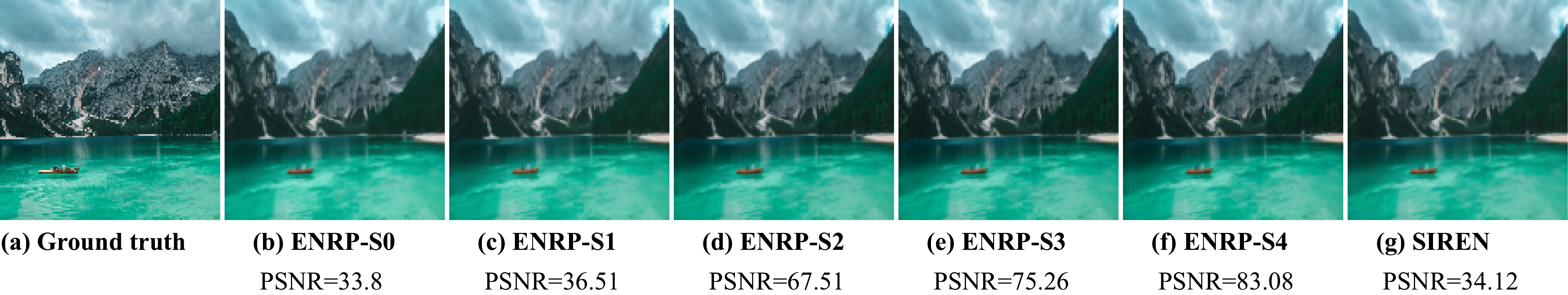}
    % \vspace{-1em}
    \vspace{-5px}
    \caption{Comparison of different implicit network architectures trained on a $128\times128$ ground truth (a) after 500 iterations.}
    \label{fig:result}
    % \vspace{-1em}
    \vspace{-10px}
\end{figure*}

% \vspace{-5px}
In our experiment, we set the $iter_{max}$ of algorithm (\hyperref[algorithm:main]{1}) equal to 5, and $F_{max}$ equal to $1$ Giga FLOPs. In order to find the best architecture in the second and third lines of the algorithm, we train the architectures with different $w$ and $d$ values from the search sets on the image presented in Fig. \ref{fig:result}. Further, to diminish the slight randomness effect of models parameters initialization, we trained each model in the algorithm process for 5 times. In Fig. \ref{fig:algorithm_plot}, we have shown the PSNR of the designed ensemble networks at each iteration of the algorithm (\hyperref[algorithm:main]{1}) for $M = \{1, 2, \dots, 32\}$. For $M=1,2, \textrm{and}$ $4$, the algorithm converges in $3$ iterations, and for $M=8, 16, \textrm{and}$ $32$, the algorithm stops after $2$ iterations.

Our proposed architectures outperform SIREN in both terms of quantitative (Table \ref{tab:table1} and Fig. \ref{fig:result_iter}) and qualitative (Fig. \ref{fig:result}). %Let us denote the ensemble neural representation networks (ENRP) by ENRP-S0 to ENRP-S4 for $M = \{2, 4, \dots, 32\}$.
In Fig. \ref{fig:result}, an image is represented by our proposed ensemble neural representation networks (ENRP) for $M = \{2, 4, \dots, 32\}$, i.e., ENRP-S0 to ENRP-S4, and SIREN.
\begin{figure}[t]
    % \vspace{-15px}
    \centering
    \includegraphics[width=7cm]{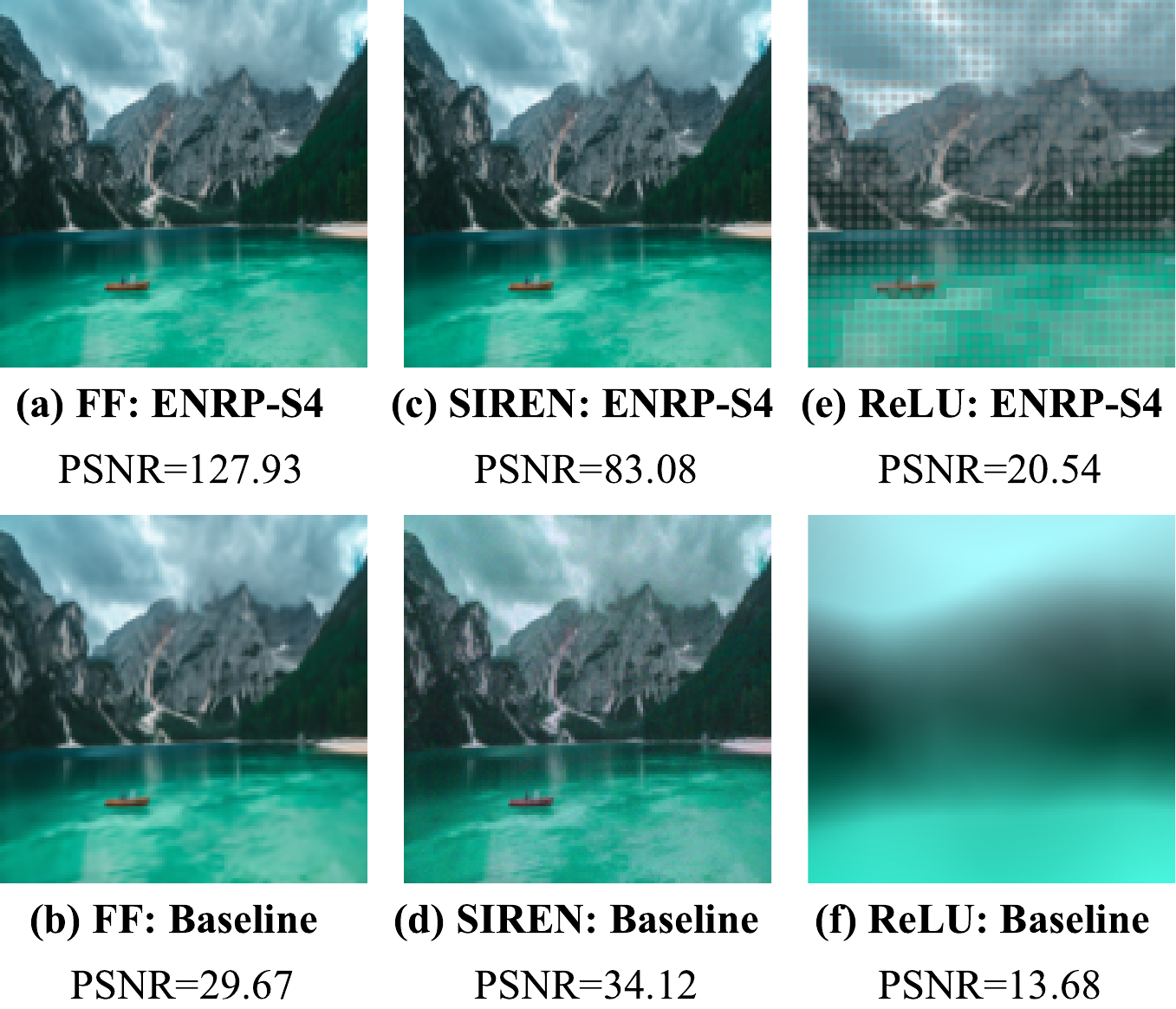}
    % \vspace{-1em}
    \vspace{-5px}
    \caption{Comparison between Fourier feature (FF), ReLU-based, and SIREN models in an approximately equal number of FLOPs after 500 iterations}
     \label{fig:comparison}
    \vspace{-0.8em}
\end{figure}

In Fig. \ref{fig:result_iter}, the PSNR versus training iterations for ENRP and SIREN models are shown. Our ENRP models from S0 to S4 generally have an order of magnitude fewer FLOPs than SIREN while attaining higher PSNR values. In particular, our ENRP-S4 achieves $83.08$ dB PSNR with $557.84$ Mega FLOPs, which requires almost $11.62$ times fewer FLOPs than SIREN, and its PSNR value has increased by $\%143.49$. Additionally, Fig. \ref{fig:result_iter} illustrates that the convergence rate of the proposed architectures is much higher compared to SIREN, and ENRP-S2 to S4 could surpass the SIREN in terms of PSNR values after the first $80$ steps.

For showing the generality of the proposed approach, we have also ran it on ReLU-based and fourier Feature networks for grid size 32 with 557.84 M FLOPs and 558.11 M FLOPs, respectively. For the Fourier feature network, the mapping size parameter was set to 65. As shown in Fig. \ref{fig:comparison}, the results of the ENRP networks saw significant improvements compared to those of their baseline networks.

Furthermore, we have expanded our approach to audio signals and ran our iterative algorithm on
% {\color{red} (specification of the audio signal?)}
audio data for three iterations with $M = \{2, 4, \dots, 32\}$, i.e., ENRP-P0 to ENRP-P4 and SIREN. In this process, we exploited the 6-second audio available at \url{https://github.com/vsitzmann/siren/blob/master/data/gt_bach.wav} with a sampling rate of 44100 \nolinebreak Hz. In the pre-processing step, we scaled the timescale of the data to be in the range of [-1, 1]. Then, we equally divided the signal into $m$ shorter parts in the time domain for $M=m$ of grid ratio. Finally, each of these shorter parts is represented by a separate sub-model. As shown in Table \ref{tab:table2}, we observed promising results such that our proposed ensemble network can achieve up to $\%14$ higher PSNR compared to that of the base SIREN model while requiring about one-tenth of FLOPS.

% This result demonstrates the efficiency of our method on audio signals, and can be extended to other types of signals, including 3D videos.

\begin{table}[t]
    \vspace{0.5em}
  \begin{center}
    \caption{\textbf{Audio:} The configurations, PSNRs, and FLOP counts of the proposed ensemble networks for $M = \{2, 4, \dots, 32\}$ and SIREN, after 500 training steps.}
    \label{tab:table2}
    \vspace{-10px}
    \begin{tabular}{c|c|c|c|c|c}
      \textbf{Name} & \textbf{M} & \textbf{Depth (d)} & \textbf{Width (w)} & \textbf{$\#$FLOPs} & \textbf{PSNR (dB)}\\
      \hline SIREN & 1 & 3 & 256 & 6.48 G & 44.56\\
      \hline ENRP-P0 & 2  & 5 & 64 &  681.58 M & 36.02 \\
      \hline ENRP-P1 & 4  & 5 & 64 &  681.58 M & 35.91 \\
      \hline ENRP-P2 & 8  & 5 & 64 &  681.58 M & 36.26 \\
      \hline ENRP-P3 & 16 & 5 & 64 &  681.58 M & 42.23 \\
      \hline ENRP-P4 & 32 & 5 & 64 &  681.58 M & 50.89
    \end{tabular}
  \end{center}
  \vspace{-20px}
    % \vspace{-3em}

\end{table}

%%%%%%%%%%%%%%%%%%%%%%%%%%%%%%%%%%%%%%%%%%%%%%%%%%%%%%%%%%%%%%%%%%%%%%%%%%%%%%%%%%%%%%%%%%%%%%%%%%%%%%%%%%%%%%%%%%%%%%%%%%%%%%%%%%%%%%%%%%%%%%%%%%%%%%%%%%%%%%%%%%%%%%%%%%%%%%%%%%%%%%%%%%%%%%%%%%%%%%%%%%%%%%%%

% \vspace{-1em}
\section{Conclusion}
\label{sec:conclusion}

In this paper, we have proposed ensemble INR networks that obtain better performance with lower FLOP counts than those of SIREN, ReLU, and Fourier feature networks. This fewer required FLOP count and the parallelized implementation result in a significant decrease in training and inference times that lets us utilize INR networks in real-world applications. We observed that increasing the width of sub-networks forming a large ensemble network does not always lead to better performance; hence, we have proposed an iterative regularized optimization algorithm that searches for the best performing architecture while satisfying a FLOP count constraint. The simulation results show the superiority of the proposed architecture over its counterpart for different natural signals.

\newpage
\bibliographystyle{IEEEtran}
\bibliography{IEEEabrv,ENR_ms}

\end{document}